\documentclass[runningheads]{llncs}
\usepackage[T1]{fontenc}
\usepackage{graphicx}
\usepackage{amsmath}
\usepackage{amssymb}
\usepackage{amsfonts}
\usepackage{booktabs}
\usepackage{xcolor}
\usepackage{multirow}
\usepackage{wrapfig}
\usepackage{tabularx}
\usepackage{adjustbox}
\usepackage{hyperref}
\usepackage{float}
\usepackage{needspace}
\usepackage{placeins}
\usepackage{amsmath}
 
\begin{document}
\title{HPMixer: Hierarchical Patching for Multivariate Time Series Forecasting}

%
\author{Jung Min Choi \and
Vijaya Krishna Yalavarthi \and
Lars Schmidt-Thieme}
%
\authorrunning{J.Choi et al.}
%
\institute{ISMLL, University of Hildesheim \\
\email{choij@uni-hildesheim.de}}
%
\maketitle              

\begin{abstract}
   In long-term multivariate time series forecasting, effectively capturing both periodic patterns and residual dynamics is essential. 
   To address this within standard deep learning benchmark settings, we propose the \textbf{H}ierarchical \textbf{P}atching \textbf{Mixer} (\textbf{HPMixer}), which models periodicity and residuals in a decoupled yet complementary manner. 
   The periodic component utilizes a learnable cycle module~\cite{cyclenet} enhanced with a nonlinear channel-wise MLP for greater expressiveness. 
   The residual component is processed through a Learnable Stationary Wavelet Transform (LSWT) to extract stable, shift-invariant frequency-domain representations. 
   Subsequently, a channel-mixing encoder models explicit inter-channel dependencies, while a two-level non-overlapping hierarchical patching mechanism captures coarse- and fine-scale residual variations. 
   By integrating decoupled periodicity modeling with structured, multi-scale residual learning, HPMixer provides an effective framework. 
   Extensive experiments on standard multivariate benchmarks demonstrate that HPMixer achieves competitive or state-of-the-art performance compared to recent baselines.
   
\end{abstract}
   
\section{Introduction}
\label{introduction}

In long-term multivariate time series (MTS) forecasting, learning periodicity is particularly crucial, as it often provides the strongest and most stable predictive signals. 
Recent works such as CycleNet \cite{cyclenet} demonstrate the effectiveness of explicitly modeling periodic components through learnable recurrent cycles, showing that a well-captured cycle structure can significantly improve forecasting accuracy over long horizons. 
However, periodicity alone cannot fully characterize real-world time series. Many datasets exhibit components that are not strictly periodic, including long-term trends, irregular fluctuations, structural changes, and elements that remain in the residuals after periodic patterns are extracted. 
These residuals are not merely noise but contain meaningful dynamics that influence future values, and failing to model them properly leads to accumulated errors, especially in long-range prediction settings. 
Therefore, it is essential to complement periodicity learning with a dedicated mechanism that can effectively capture and represent residual behavior, rather than treating it as an unstructured remainder.

Our proposed model, \textbf{H}ierachcial \textbf{P}atching \textbf{Mixer} (\textbf{HPMixer}), is a novel architecture that models temporal recurrence in the periodic component, while the residual component captures rich temporal and channel interdependencies.
It leverages a hierarchical, non-overlapping patching mechanism and a learnable Stationary Wavelet Transform (SWT). 
The architecture also integrates learnable cycle modules from CycleNet \cite{cyclenet}, enhanced with an MLP module that refines the periodic representation across channels.
This deepens the periodic modeling capacity, enabling the module to refine and express more intricate periodic behaviour.
Additionally, the hierarchical non-overlapping patching enables the model to learn detailed temporal information, capturing relationships between both close and distant time steps. 
This approach also allows the model to focus on channel relationships between neighboring time steps, as well as potential channel dependencies that may occur with some time lags.
The SWT is particularly well suited for time series forecasting because it preserves temporal resolution across all scales and provides a stable multi-resolution representation without information loss. 
When combined with learnable kernel filters \cite{leranableswt}, SWT can adaptively capture meaningful structures in the data, offering improved representation of both periodic and non-periodic components as well as stronger denoising capability. 
In our model, we incorporate a learnable SWT module to exploit these advantages for more expressive residual modeling.
The \textbf{main contributions} of our paper are:

\begin{enumerate}
   \item[(i)] We show that a hierarchical non-overlapping patching mechanism is effective for capturing both channel and temporal relationships.
   \item[(ii)] We demonstrate that learning on both periodicity and residuals is crucial for handling complex, high-dimensional datasets and for long-term forecasting. 
   \item[(iii)] Finally, we propose \textbf{HPMixer}, which achieves state-of-the-art results across various datasets. Code available at: \href{https://github.com/choijm-p/HPMixer}{https://github.com/choijm-p/HPMixer}. 
\end{enumerate}

\section{Related Works}
\label{related_works}

\subsection{Patching in Time Series Forecasting}

The patching mechanism is originally from the Vision Transformer (ViT), where it was introduced to enable Transformer models to process image data by treating an image as a sequence of smaller patches \cite{vit}.
It allows the self-attention mechanism to scale to high-resolution images by reducing the number of tokens it has to process, enabling the model to capture long-range dependencies across the entire image effciently.

PatchTST \cite{patchtst} introduces overlapping patches for MTS forecasting to capture local semantics.
TSMixer \cite{tsmixer2} and WPMixer \cite{wpmixer} also adopt this approach, using MLPs to learn temporal and channel dependencies.
Overlapping patches help preserve information across patch boundaries, ensuring smoother and more complete temporal representations.

Our paper introduces a non-overlapping patching mechanism, similar to ViT, achieving strong performance with improved computational efficiency and reduced redundancy. The Swin Transformer \cite{swintransformer} employs a hierarchical patching strategy that begins with small non-overlapping patches and progressively merges them in deeper layers, capturing both fine local details and global context with linear complexity. Inspired by this, our model adopts a two-level hierarchical patching mechanism to learn interdependencies across channels and capture semantic information over both nearby and distant time steps.
\subsection{Frequency Domain in Times Series Forecasting}

Many models transform time-domain data into the frequency domain to reveal periodic and cyclical patterns that are hard to detect in raw signals.
Fedformer \cite{fedformer} applies Discrete Fourier Transform (DFT) to obtain frequency spectra for self-attention, while Autoformer \cite{autoformer} replaces self-attention with an Fast Fourier Transform (FFT)-based auto-correlation mechanism.
FreTS \cite{frets} and MSGNet \cite{msgnet} also use FFT to capture long-range dependencies and identify key multi-periodicities across channels.
WaveForM \cite{waveform} and WPMixer \cite{wpmixer} employ the Discrete Wavelet Transform (DWT) to capture both short-term fluctuations and long-term trends.
However, DWT is not shift-invariant, a small shift in input can produce completely different coefficients due to its downsampling process, which may miss important features.
To address this, the Learnable Stationary Wavelet Transform (LSWT) avoids downsampling and computes coefficients at every possible shift \cite{swt}.
With learnable kernel filters \cite{leranableswt}, LSWT can adaptively model complex periodicities and trends with superior denoising capability.
Our model employs the LSWT to fully leverage these advantages.

\subsection{Channel Dependence}
Channel-independent models process each channel as a separate univariate series, ignoring inter-channel dependencies, while channel-dependent models explicitly learn these relationships.
PatchTST \cite{patchtst} and CycleNet \cite{cyclenet} focus on temporal patterns within each channel, whereas iTransformer \cite{itransformer} learns only cross-channel dependencies via self-attention.
Relying solely on either approach makes performance dataset-dependent.
To address this, models such as Crossformer \cite{crossformer}, TSMixer \cite{tsmixer}, and TimeMixer \cite{timemixer} combine univariate and multivariate components to jointly capture both temporal and inter-channel correlations.
Our model likewise leverages both channel dependence and channel independence in order to learn the underlying dynamics more effectively.
\section{Proposed Methodology}

\subsection{Preliminaries}
A MTS is a collection of $C$ variables observed over time. 
It can be represented as a sequence of data points $X = \{x_{t-L+1}, \dots, x_t\}$, where each vector $x_t \in \mathbb{R}^C$ contains the observations from all channels at time step $t$. 
This sequence forms a matrix $X \in \mathbb{R}^{C \times L}$, where $L$ is the look-back window.
The task of MTS forecasting is to learn a model that can predict the future values of these $C$ variables based on their past observations. Given an input time series $X$, the objective is to forecast the subsequent $H$ time steps, which we represent as a matrix $X_H \in \mathbb{R}^{C \times H}$, where $H$ is the forecasting horizon.
From a training dataset $D_{train}$, we aim to find an optimal model $\hat{y}$ that approximates a function mapping the input series to the predicted output series, $\hat{y}: \mathbb{R}^{C \times L} \to \mathbb{R}^{C \times H}$.

\subsection{Model Architecture}

\begin{figure}[htbp]
   \begin{center}
   \includegraphics[width=1\textwidth]{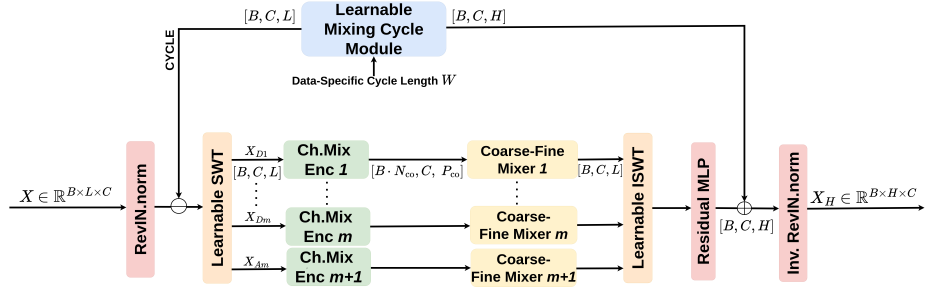}
   \end{center}
   \caption{An overview of the proposed model HPMixer architecture.}
   \label{fig:model_architecture}
\end{figure}

The overall architecture of the proposed model is illustrated in Fig.~\ref{fig:model_architecture}.
At a high level, the model processes periodicity and residual components through separate branches.

In the periodicity branch, the input is passed through a Learnable Mixing Cycle Module, which extends the Recurrent Cycle Module from \cite{cyclenet}.
While CycleNet learns cycles independently for each channel and does not deepen the periodic modeling beyond simple repetition, our model extends it with an MLP that enriches the periodic representation.

In the residual branch, the input is first decomposed by the LSWT.
The resulting coefficients are going through a hierarchical patching mechanism, forming coarse- and fine-patches.
As shown in Fig.~\ref{fig:mixerblock}, these patches are processed by a Channel-Mixing Encoder and Coarse-Fine Patching Mixer to learn intra- and inter-patch correlations.

After the mixing stages, the features are reconstructed to the time domain using an Inverse Stationary Wavelet Transform (ISWT), followed by an MLP to capture global temporal dependencies.
Finally, the residual branch output is combined with the output of the periodicity branch to produce the final prediction.

\subsubsection{Learnable Mixing Cycle Module}

\begin{figure}[htbp]
   \centering
   \includegraphics[width=0.8\textwidth]{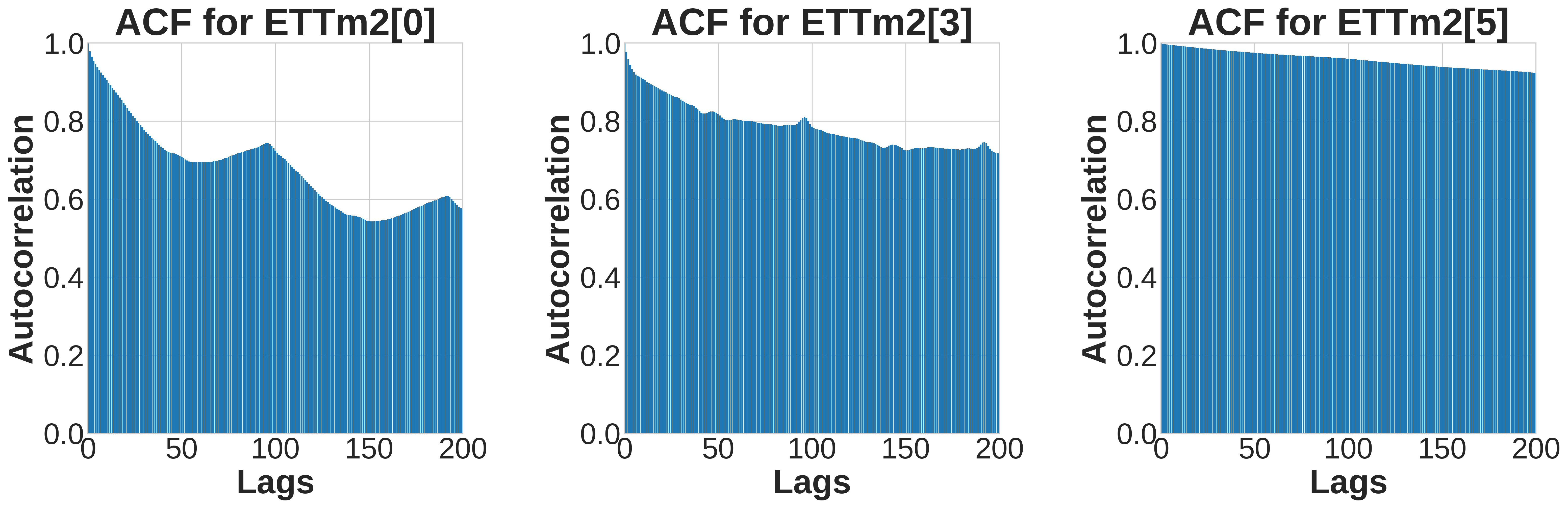}
   \caption{Autocorrelation (ACF) plots for three channels of the ETTm2 dataset. 
The first channel (ETTm2[0]) exhibits a clear periodic signal with noticeable peaks every 96 time steps, 
whereas the third channel (ETTm2[3]) shows a weaker and less regular periodic pattern. 
The fifth channel (ETTm2[5]) displays only very vague cyclic structure at the 96-step frequency.}
   \label{fig:acf_fig}
\end{figure}

\begin{figure}[htbp]
   \begin{center}
   \includegraphics[width=0.9\textwidth]{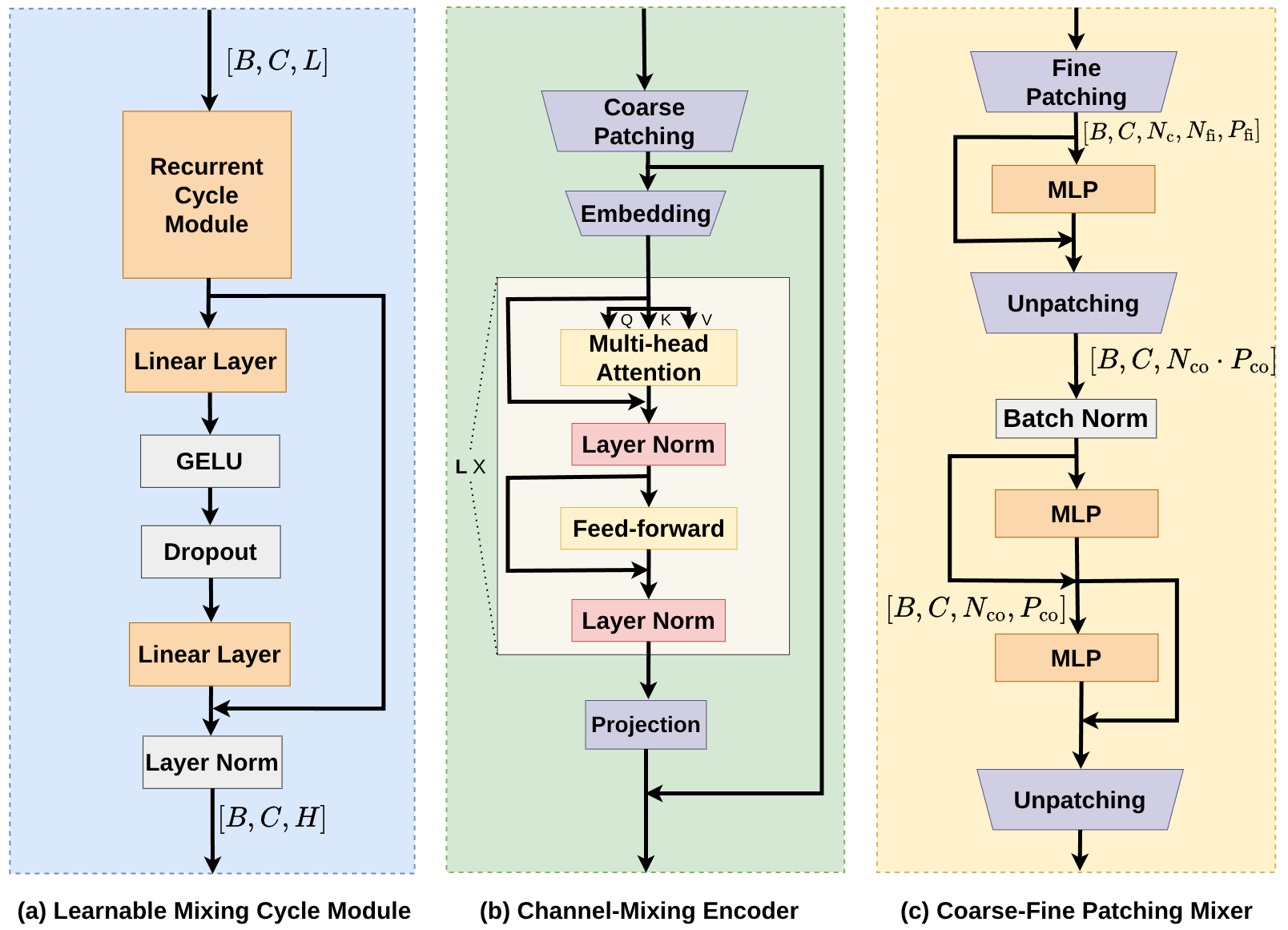}
   \end{center}
   \caption{Architecture of the \textit{Learnable Mixing Cycle Module}, the \textit{Channel-Mixing Encoder}, and \textit{Coarse-Fine Patching Mixer}.}
   \label{fig:mixerblock}
\end{figure}

The recurrent cycle module from \cite{cyclenet} learns periodic patterns by learning a set of recurrent cycles via gradient-based training. 
It constructs a parameter matrix $Q \in \mathbb{R}^{W \times C}$, where $W$ is a preset cycle length and $C$ is the number of channels. 
To ensure that $W$ captures the correct intrinsic periodicity, we follow the methodology of CycleNet and determine its value prior to training by analyzing the Autocorrelation Function (ACF) of the training data, selecting the time lag that corresponds to the highest dominant peak.
Each channel is assigned a learned cycle represented by one column of $Q$, and the model extracts cyclic components from the input sequence through alignment and repetition.
Since the cycle module learns each feature’s periodicity independently and does not deepen periodic learning through any nonlinear refinement, it can struggle on multivariate datasets where periodic patterns differ or are weak across features.

As illustrated in Fig.~\ref{fig:acf_fig}, the ACF patterns of ETTm2 reveal substantial variation in periodic behavior across channels. 
While the first channel (ETTm2[0]) shows strong and well-aligned peaks at multiples of the 96-step cycle, the third channel (ETTm2[3]) exhibits only moderate and less stable periodic structure, and the fifth channel (ETTm2[5]) shows almost no discernible peaks at this interval. 
This indicates that a single predefined cycle length cannot adequately represent heterogeneous or weak periodic patterns.

To address this limitation, we enhance the learnable cycle module with an MLP that increases the model’s capacity to learn periodic structure. 
Although the module does not directly receive the input sequence~$X$, the MLP introduces additional channel-wise processing that enriches the learned periodic representation. 
Given an initial periodic representation $X_{\text{period}} \in \mathbb{R}^{W \times C}$, the refined periodic component is computed as
\begin{equation}
   \hat{X}_{\text{period}} 
   = \mathit{LayerNorm}\!\left(\mathit{MLP}(X_{\text{period}})\right)
   \quad \in \mathbb{R}^{W \times C}.
\end{equation}
This refinement allows the periodic module to form a more expressive across channels, which helps accommodate channels with weak or heterogeneous periodicities, as observed in the ACF analysis. 
The structure of this module is illustrated in Figure~\ref{fig:mixerblock}(a).
   
\subsubsection{Learnable Stationary Wavelet Transform}
The SWT is a time-invariant multi-scale decomposition in which a signal is convolved with dilated low-pass ($h_0^{(j)}$) and high-pass ($h_1^{(j)}$) analysis filters at each level $j$. This decomposition yields a single approximation coefficient $A_k$ at the highest level, alongside $k$ detail coefficients $D_1, \dots, D_k$.
\begin{align}
(A_j, D_j)
&= \mathcal{W}_j(X)
= \bigl( X * h_0^{(j)},\; X * h_1^{(j)} \bigr),
\qquad j = 1, \dots, k, \\[4pt]
X
&= \mathcal{W}^{-1}(A_k, D_1, \dots, D_k)
= A_k * g_0^{(k)}
\;+\;
\sum_{j=1}^{k} D_j * g_1^{(j)} ,
\end{align}
where $h_0^{(j)}$ and $h_1^{(j)}$ are the $j$-level dilated analysis filters, and $g_0^{(j)}$ and $g_1^{(j)}$ are their corresponding synthesis filters. Learning representations in the frequency domain is beneficial for multivariate time series forecasting because periodic structure and cross-channel dependencies often become more separable and easier to model in the spectral domain than in the raw time domain.
The LSWT replaces the fixed wavelet filters with trainable parameters, allowing the model to learn task-specific low-pass and high-pass filters that better separate trends from high-frequency variations. 
This produces more informative decompositions than predefined wavelets and improves downstream prediction performance. 

\subsubsection{Hierarchical Non-Overlapping Patching}
The decomposed residual tensors $D_m, A_m \in \mathbb{R}^{C \times L}$ are processed using a two level non-overlapping patching strategy.
The first level, coarse patching, divides the sequence into $N_{\text{co}}$ patches of length $P_{\text{co}}$, forming a tensor $X_{\text{coarse}} \in \mathbb{R}^{C \times P_{\text{co}} \times N_{\text{co}}}$.
In the second level, fine-patching, each coarse patch is further split into $N_{\text{fi}}$ smaller patches of length $P_{\text{fi}}$, producing $X_{\text{fine}} \in \mathbb{R}^{C \times N_{\text{co}} \times P_{\text{fi}} \times N_{\text{fi}}}$.
For simplicity, the approximation and detail components ($D_m$, $A_m$) are jointly written as $X_d$.

\begin{enumerate}
   \item \textbf{Coarse patching:}
   \[
   X_{\text{coarse}} = \mathit{Patch}_{co}(X_d), 
   \qquad 
   X_{\text{coarse}} \in \mathbb{R}^{C \times P_{\text{co}} \times N_{\text{co}}},
   \qquad 
   N_{\text{co}} = \left\lfloor \frac{L}{P_{\text{co}}} \right\rfloor 
   \] 
   \item \textbf{Fine patching:}
   \[
   X_{\text{fine}, n} = \mathit{Patch}_{fi}(X_{\text{coarse}, n}), 
   \qquad 
   X_{\text{fine}, n} \in \mathbb{R}^{C \times P_{\text{fi}} \times N_{\text{fi}}},
   \qquad 
   N_{\text{fi}} = \left\lfloor \frac{P_{\text{co}}}{P_{\text{fi}}} \right\rfloor 
   \]
\end{enumerate}

This hierarchical structure allows the model to learn localized short range patterns from fine patches while still preserving long range temporal context through the coarse patches.

\subsubsection{Channel Mixing Encoder}
After coarse patching, the Channel Mixing Encoder is applied to each patch in order to capture cross channel-dependencies before the fine scale processing stage.
The encoder adopts a Transformer \cite{attention} inspired structure that contains multi-head attention, layer normalization, a feed forward network, and residual connections.
It mixes information across channels so that shared temporal patterns can be learned, instead of processing each variable in isolation.
The overall structure of this encoder is illustrated in Figure~\ref{fig:mixerblock}(b).

\subsubsection{Coarse-Fine Patching Mixer}
The output of the Channel Mixing Encoder, $X_d \in \mathbb{R}^{C \times N_{\text{co}} \times P_{\text{co}}}$, is refined by a two stage mixer that operates at the coarse and fine patch levels.
Since the LSWT produces one approximation coefficient and $m$ detail coefficients, the mixer contains $(m+1)$ independent MLP blocks, one for each coefficient branch. The structure of this mixer is presented in Figure~\ref{fig:mixerblock}(c).

\paragraph{Fine-Patch Mixer}
A fine scale MLP mixes information across the $N_{\text{fi}}$ fine patches for every coefficient branch:
\begin{equation}
   \hat{X}_{\text{fine}} = \mathit{MLP}_{fi}(X_{\text{fine}}) + X_{\text{fine}}
   \qquad \in \mathbb{R}^{C \times N_{\text{co}} \times N_{\text{fi}} \times P_{\text{fi}}} .
\end{equation}

\paragraph{Coarse-Patch Mixer}
The fine scale output is first mixed at the expanded coarse resolution of length $P_{\text{co}} N_{\text{co}}$ and then refined again at the coarse length $P_{\text{co}}$:
\begin{equation}
   \tilde{X}_{\text{coarse}} = \mathit{MLP}_{\text{flat}}(\hat{X}_{\text{fine}}) + \hat{X}_{\text{fine}}
   \qquad \in \mathbb{R}^{C \times (N_{\text{co}} P_{\text{co}})}
   \end{equation}
   
   \begin{equation}
   \hat{X}_{\text{coarse}} = \mathit{MLP}_{\text{patch}}(\tilde{X}_{\text{coarse}}) + \tilde{X}_{\text{coarse}}
   \qquad \in \mathbb{R}^{C \times N_{\text{co}} \times P_{\text{co}}}
\end{equation}
   
This two stage mixing process first blends fine scale information within each coarse window at the expanded temporal resolution, and then refines the representation at the original coarse scale. 
This lets the model learn local variations inside each segment while integrating information across patches to characterize extended temporal dynamics.

\subsubsection{Prediction Layer}
The output of the ISWT, denoted as $\hat{X}_{\text{recon}} \in \mathbb{R}^{C \times L}$, is fed into an MLP with a residual connection to learn temporal dependencies across the entire sequence length.
This is then passed to a final linear prediction layer to produce the forecast, which is subsequently added to the predicted periodicity to form the final output $\hat{Y} \in \mathbb{R}^{C \times H}$.
\begin{equation}
   \hat{Y} = 
   \mathit{Linear}\big(
   \mathit{MLP}_{\text{res}}(\hat{X}_{\text{recon}}) + \hat{X}_{\text{recon}}
   \big)
   + \hat{Y}_{\text{period}}
   \quad \in \mathbb{R}^{C \times H}
\end{equation}

\section{Experiments}
\subsection{Experimental Setup}
The experiments are conducted using seven widely adopted benchmark datasets: ETTh1, ETTh2, ETTm1, ETTm2, Weather, Electricity (ECL), and Traffic~\cite{informer,autoformer}.

The model is trained to forecast over four distinct long-term horizons: 96, 192, 336, and 720. The training objective is to minimize the Mean Squared Error (MSE), while the performance on the test set is evaluated using both the MSE and the Mean Absolute Error (MAE).
We compare our model with six recent state-of-the-art models: SimpleTM~\cite{simpletm}, CycleNet~\cite{cyclenet}, Timexer~\cite{timexer}, iTransformer~\cite{itransformer}, PatchTST~\cite{patchtst}, and TimeMixer~\cite{timemixer}. 
To ensure full reproducibility, the complete source code and exact hyperparameter configurations for all datasets are available in the anonymous repository linked in Section~\ref{introduction}.
\subsection{Results}

\begin{table}[!htbp]
\caption{Full results over different horizons on the standard datasets. The results are an average of 5 different random seeds \{3000, 3001, 3002, 3003, 3004\}. 
The best variant for each model is in \textbf{bold} and the second best model is in \underline{{Underline}}. The MSE/MAE values for the baselines are imported from \cite{simpletm}, \cite{itransformer}, \cite{cyclenet}, and \cite{timemixer}.}
\label{tab:Full-results-standard-datasets}
\begin{center}
\LARGE
\setlength{\tabcolsep}{6pt}
\renewcommand{\arraystretch}{1.2}
\begin{adjustbox}{max width=\textwidth}
\begin{tabular}{l|c|cc|cc|cc|cc|cc|cc|cc}
\toprule
Dataset & Horizon & \multicolumn{2}{c}{HPMixer (Ours)} & \multicolumn{2}{c}{SimpleTM} & \multicolumn{2}{c}{CycleNet/MLP} & \multicolumn{2}{c}{TimeXer} & \multicolumn{2}{c}{iTransformer} & \multicolumn{2}{c}{PatchTST} & \multicolumn{2}{c}{TimeMixer} \\
\midrule
   \multicolumn{2}{c|}{} & MSE & MAE & MSE & MAE & MSE & MAE & MSE & MAE & MSE & MAE & MSE & MAE & MSE & MAE \\
\midrule
\multirow{5}{*}{ETTm1} & 96 & \textbf{0.305} & \textbf{0.352} & 0.321 & 0.361 & 0.319 & 0.360 & \underline{0.318} & \underline{0.356} & 0.334 & 0.368 & 0.329 & 0.367 & 0.328 & 0.363 \\
& 192 & \textbf{0.345} & \textbf{0.377} & \underline{0.360} & \underline{0.380} & \underline{0.360} & 0.381 & 0.362 & 0.383 & 0.377 & 0.391 & 0.367 & 0.385 & 0.364 & 0.384 \\
& 336 & \textbf{0.375} & \textbf{0.398} & 0.390 & 0.404 & \underline{0.389} & \underline{0.403} & 0.395 & 0.407 & 0.426 & 0.420 & 0.399 & 0.410 & 0.390 & 0.404 \\
& 720 & \textbf{0.432} & \textbf{0.433} & 0.454 & \underline{0.438} & \underline{0.447} & 0.441 & 0.452 & 0.441 & 0.491 & 0.459 & 0.454 & 0.439 & 0.458 & 0.445 \\
\cline{2-16}
& Avg  & \textbf{0.364}  & \textbf{0.390} & 0.381 & \underline{0.396} & \underline{0.379} & \underline{0.396} & 0.382 & 0.397 & 0.407 & 0.410 & 0.387 & 0.400 & 0.385 & 0.399 \\
\midrule
\multirow{5}{*}{ETTm2} & 96 & \textbf{0.160} & \textbf{0.228} & 0.173 & 0.257 & \underline{0.163} & \underline{0.246} & 0.171 & 0.256 & 0.180 & 0.264 & 0.175 & 0.259 & 0.176 & 0.259 \\
& 192 & \textbf{0.224} & \textbf{0.286} & 0.238 & 0.299 & \underline{0.229} & \underline{0.290} & 0.237 & 0.299 & 0.250 & 0.309 & 0.241 & 0.302 & 0.242 & 0.303 \\
& 336 & \textbf{0.280} & \textbf{0.323} & 0.296 & 0.338 & \underline{0.284} & \underline{0.327} & 0.296 & 0.338 & 0.311 & 0.348 & 0.305 & 0.343 & 0.304 & 0.342 \\
& 720 & \textbf{0.383} & \textbf{0.388} & 0.393 & 0.395 & 0.389 & \underline{0.391} & \underline{0.392} & 0.394 & 0.412 & 0.407 & 0.402  & 0.400 & 0.393 & 0.397 \\
\cline{2-16}
& Avg & \textbf{0.261} & \textbf{0.306} & 0.278 & 0.325 & \underline{0.266} & \underline{0.314} & 0.274 & 0.322 & 0.288 & 0.332 & 0.281 & 0.326 & 0.278 & 0.325 \\
\midrule
\multirow{5}{*}{ETTh1} & 96 & \underline{0.371} & \underline{0.395} & \textbf{0.366} & \textbf{0.392} & 0.375 & 0.395 & 0.382 & 0.403 & 0.386 & 0.405 & 0.414 & 0.419 & 0.381 & 0.401 \\
& 192 & \textbf{0.416} & \underline{0.423} & \underline{0.422} & \textbf{0.421} & 0.436 & 0.428 & 0.429 & 0.435 & 0.441 & 0.436 & 0.460 & 0.445 & 0.440 & 0.433  \\
& 336 & \underline{0.453} &\underline{0.445} & \textbf{0.440} & \textbf{0.438} & 0.496& 0.455 & 0.468 & 0.448 & 0.487 & 0.458 & 0.501 & 0.466 & 0.501 & 0.462  \\
& 720 & \textbf{0.455} & \textbf{0.460} & \underline{0.463} & \underline{0.462} & 0.520 & 0.484 & 0.469 & 0.461 & 0.503 & 0.491 & 0.500 & 0.488 & 0.501 & 0.482 \\
\cline{2-16}
& Avg & \underline{0.423} & \underline{0.430} & \textbf{0.422} & \textbf{0.428} & 0.457 & 0.441 & 0.437 & 0.437 & 0.454 & 0.447 & 0.469 & 0.454 & 0.458 & 0.445  \\
\midrule
\multirow{5}{*}{ETTh2} & 96 & \textbf{0.280} & \textbf{0.333} & \underline{0.281} & \underline{0.338} & 0.298 & 0.344 & 0.286 & 0.338 & 0.297 & 0.349 & 0.302 & 0.348 & 0.292 & 0.343 \\
& 192 & \underline{0.363} & \textbf{0.386} & \textbf{0.355} & \underline{0.387} & 0.372 & 0.396 & 0.363 & 0.389 & 0.380 & 0.400 & 0.388 & 0.400 & 0.374 & 0.395 \\
& 336 & \underline{0.409} & \underline{0.425} & \textbf{0.365} & \textbf{0.401} & 0.431 & 0.439 & 0.414 & 0.423 & 0.428 & 0.432 & 0.426 & 0.433 & 0.428 & 0.433 \\
& 720 & 0.418 & 0.437 & \underline{0.413} & \underline{0.436} & 0.450 & 0.458 & \textbf{0.408} & \textbf{0.432} & 0.427 & 0.445 & 0.431 & 0.446 & 0.454 & 0.458 \\
\cline{2-16}
& Avg & \underline{0.367} & \underline{0.402} & \textbf{0.353} & \textbf{0.391} & 0.388 & 0.409 & 0.367 & 0.396 & 0.383 & 0.407 & 0.387 & 0.407 & 0.384 & 0.407 \\
\midrule
\multirow{5}{*}{Electricity} & 96 & \textbf{0.133} & \textbf{0.230} & 0.141 & \underline{0.235} & \underline{0.136} & \textbf{0.229} & 0.140 & 0.242 & 0.148 & 0.240 & 0.181 & 0.270 & 0.153 & 0.244 \\
& 192 & \textbf{0.151} & 0.248 & \textbf{0.151} & \underline{0.247} & \underline{0.152} & \textbf{0.244} & 0.157 & 0.256 & 0.162 & 0.253 & 0.188 & 0.274 & 0.166 & 0.256 \\
& 336 & \textbf{0.169} & \textbf{0.264} & 0.173 & 0.267 & \underline{0.170} & \textbf{0.264} & 0.176 & 0.275 & 0.178 & 0.269 & 0.204 & 0.293 & 0.184 & 0.275  \\
& 720 & \textbf{0.201} & \underline{0.296} & \textbf{0.201} & \textbf{0.293} & 0.212 & 0.299 & \underline{0.211} & 0.306 & 0.225 & 0.317 & 0.246 & 0.324 & 0.226 & 0.313 \\
\cline{2-16}
& Avg & \textbf{0.163} & \textbf{0.259} & \underline{0.166} & \underline{0.26} & 0.168 &  0.259 & 0.171 & 0.270 & 0.178 & 0.270 & 0.205 & 0.290 & 0.182 & 0.272 \\
\midrule
\multirow{5}{*}{Traffic} & 96 & 0.432 & \underline{0.274} & \underline{0.410} & \underline{0.274} & 0.458 & 0.296 & 0.428 & 0.271 & \textbf{0.395} & \textbf{0.268} & 0.462 & 0.295 & 0.464 & 0.289 \\
& 192 & 0.445 & \underline{0.276} & \underline{0.430} & 0.280 & 0.457 & 0.294 & 0.448 & 0.282 & \textbf{0.417} & \textbf{0.276} & 0.466 & 0.296 & 0.477 & 0.292 \\
& 336 & 0.472 & 0.298 & \underline{0.449} & \underline{0.290} & 0.470 & 0.299 & 0.473 & 0.289 & \textbf{0.433} & \textbf{0.283} & 0.482 & 0.304 & 0.500 & 0.305 \\
& 720 & 0.494 & \underline{0.301} & \underline{0.486} & 0.309 & 0.502 & 0.314 & 0.516 & 0.307 & \textbf{0.467} & \textbf{0.302} & 0.514 & 0.322 & 0.548 & 0.313 \\
\cline{2-16}
& Avg & 0.46 & \underline{0.287} & \underline{0.444} & 0.289 & 0.472 & 0.301 & 0.466 & 0.287 & \textbf{0.428} & \textbf{0.282} & 0.481 & 0.304 & 0.484 & 0.297 \\
\midrule
\multirow{5}{*}{Weather} & 96 & \textbf{0.154} & \textbf{0.202} & 0.162 & 0.207 & \underline{0.158} & \underline{0.203} & 0.157 & 0.205 & 0.174 & 0.214 & 0.177 & 0.218 & 0.165 & 0.212 \\
& 192 & \textbf{0.201} & \textbf{0.245} & 0.208 & 0.248 & 0.207 & \underline{0.247} & \underline{0.204} & \underline{0.247} & 0.221 & 0.254 & 0.225 & 0.259 & 0.209 & 0.253 \\
& 336 & \textbf{0.257} & \textbf{0.287} & 0.263 & 0.290 & 0.262 & 0.289  & \underline{0.261} & \underline{0.290} & 0.278 & 0.296 & 0.278 & 0.297 & 0.264 & 0.293 \\
& 720 & \textbf{0.340} & \textbf{0.341} & \textbf{0.340} & \textbf{0.341} & 0.344 & 0.344  & \textbf{0.340} & \textbf{0.341} & 0.358 & 0.347 & 0.354 & 0.348 & 0.342 & 0.345 \\
\cline{2-16}
& Avg & \textbf{0.238} & \textbf{0.269} & 0.243 & \underline{0.271} & 0.243 & \underline{0.271} & \underline{0.241} & \underline{0.271} & 0.243 & 0.271 & 0.259 & 0.281 & 0.245 & 0.276 \\
\specialrule{1pt}{0pt}{0pt}
\textbf{Win Count} &  & \textbf{23} & \textbf{21} & \underline{9} & \underline{8} &0 & 3 &2 & 2 &5& 5 &0 & 0 &0 & 0 \\
\bottomrule
\end{tabular}
\end{adjustbox}
\end{center}
\end{table}

Table~\ref{tab:Full-results-standard-datasets} summarizes the performance of HPMixer against recent state-of-the-art models on long-term MTS forecasting.
 Overall, HPMixer achieves the best results on most datasets and forecasting horizons, showing particularly strong performance on ETTm1, ETTm2, and Weather across all horizons.
 Compared to SimpleTM and CycleNet, HPMixer improves the average MSE on ETTm1 by 4.46\% and 3.96\%, respectively. 
In addition, the Win Count confirms this trend. 
HPMixer achieves the highest number of first-place scores (23 in MSE and 21 in MAE), far exceeding SimpleTM, CycleNet/MLP, and other baselines. 
This indicates that beyond strong average performance, HPMixer consistently ranks among the top models across a wide range of datasets and forecasting horizons.

 HPMixer also outperforms or matches SimpleTM on many datasets, indicating that hierarchical patching remains effective even when forecasting is performed in the frequency domain using LSWT.
 By structuring the sequence into coarse- and fine-grained patches, HPMixer is able to better capture frequency-domain patterns and long-range temporal dependencies.
 For the ETTh1 and ETTh2 datasets, although results are mixed, HPMixer remains competitive as outperforming weaker baselines and performing comparably to the stronger ones.
 
 On the Traffic dataset, while SimpleTM generally achieves slightly better results, most MAE values still favor HPMixer.
 As observed in iTransformer~\cite{itransformer}, the Traffic dataset benefits more from deeper, high-dimensional attention due to its complex spatiotemporal structure.
 Nevertheless, HPMixer achieves stronger predictions than many other baseline models.

 \section{Ablation Study}

 \begin{table}[!htbp]
   \centering
   \LARGE
   \caption{Ablation study results on multiple datasets. Lower MSE and MAE indicate better performance. Hyperparameters were independently optimized for each ablation setting.}
   \resizebox{\textwidth}{!}{
   \begin{tabular}{l|cc|cc|cc|cc}
   \hline
   \textbf{Model} & 
   \multicolumn{2}{c|}{\textbf{ETTm1}} & 
   \multicolumn{2}{c|}{\textbf{ETTh1}} & 
   \multicolumn{2}{c|}{\textbf{Weather}} & 
   \multicolumn{2}{c}{\textbf{ECL}} \\
   \cline{2-9}
    & MSE & MAE & MSE & MAE & MSE & MAE & MSE & MAE \\
   \hline
   w/o MLP in Mixing Cycle Module & \underline{0.365} & \textbf{0.388} & 0.442 & 0.438 & 0.248 & 0.275 & 0.166 & 0.262 \\
   w/o Mixing Cycle Module        & 0.406 & 0.414 & 0.437 & 0.434 & 0.247 & 0.277 & 0.184 & 0.278 \\
   w/o trainable SWT                      & 0.417 & 0.421 & \underline{0.426} & \underline{0.431} & 0.247 & 0.277 & \textbf{0.162} & \textbf{0.258} \\
   w/o SWT                                & 0.439 & 0.43 & 0.455 & 0.447 & 0.247 & 0.276 & \underline{0.163} & 0.26\\
   One-level Patching                     & 0.39 & 0.402 & 0.439 & 0.44 & 0.247 & 0.276 & 0.164 & 0.262 \\
   \noalign{\hrule height 1.2pt}
   CycleNet/MLP                           & 0.447 & 0.441 & 0.457 & 0.441 & \underline{0.243} & \underline{0.271} & 0.168 & 0.259 \\
   \textbf{HPMixer}                       & \textbf{0.364} & \underline{0.390} & \textbf{0.423} & \textbf{0.367} & \textbf{0.238} & \textbf{0.269} & \underline{0.163} & \underline{0.259} \\
   \hline
   \end{tabular}
   }
   \label{tab:ablation_results}
\end{table}

 \paragraph{Learnable Mixing Cycle Module}
 As shown in Table~\ref{tab:ablation_results}, both the module itself and its internal MLP contribute notably to performance.
 Removing the MLP component (\textit{w/o MLP in Mixing Cycle Module}) leads to a moderate degradation , especially in the ETTh1 and Weather datasets.
 Compared to CycleNet, the Mixing Cycle Module achieves improvements on most datasets, indicating that a more expressive and better-parameterized periodic module enhances cycle modeling.
 Eliminating the entire module (\textit{w/o Mixing Cycle Module}) results in the larger performance degradation.
 This demonstrates that the ability to learn periodicities with reduced noise and inter-channel correlation is critical for accurate long-term forecasting.
 
 \paragraph{Learnable SWT}
 When the SWT kernel parameters are fixed during training (\textit{w/o trainable SWT}), performance consistently worsens across most of the datasets except for the Electricity dataset.
 Training without SWT altogether (\textit{w/o SWT}) yields even larger errors, particularly on the ETTm1 and ETTh1 datasets.
 These findings confirm that frequency-domain learning facilitates trend extraction and denoising, providing more stable representations than raw time-domain signals.
 The results also highlight that learnable filters, rather than fixed wavelet kernels, enable the model to better adapt to dataset-specific spectral characteristics.
 
 \paragraph{One-level Patching}
 Using a single level patching mechanism instead of the proposed two level hierarchical patching leads to a consistent drop in accuracy across datasets.
 As shown in Table~\ref{tab:ablation_results}, the two-level patching improves average performance indicating that hierarchical decomposition enables the model to jointly capture fine grained short term variations and long range temporal dependencies.
 This structure allows the model to represent multi scale temporal features more effectively than simpler patching schemes.

\section{Conclusion and Future Work}

This paper introduces HPMixer for long-term multivariate time series forecasting, integrating three core components: an MLP-enhanced learnable cycle module for expressive periodic modeling, a Learnable Stationary Wavelet Transform (LSWT) for stable frequency-domain representation, and a two-level hierarchical patching mechanism for multi-scale temporal dynamics. 
While empirical results demonstrate competitive performance across standard benchmarks, the architecture has notable limitations. 
First, relying on a predetermined cycle length limits the model's adaptability to highly irregular or shifting periodicities. 
Second, performance gains diminish on massively high-dimensional datasets like Traffic, suggesting the current channel-mixing mechanism may struggle to fully capture highly complex inter-channel dependencies. 
Future work will explore fully adaptive, learnable cycle lengths and more scalable cross-channel modeling frameworks to better handle high-dimensional environments.

\bibliography{mybibliography}

@inproceedings{informer,
  title={Informer: Beyond efficient transformer for long sequence time-series forecasting},
  author={Zhou, Haoyi and Zhang, Shanghang and Peng, Jieqi and Zhang, Shuai and Li, Jianxin and Xiong, Hui and Zhang, Wancai},
  booktitle={AAAI},
  year={2021}
}

@article{attention,
  title={Attention is all you need},
  author={Vaswani, Ashish and Shazeer, Noam and Parmar, Niki and Uszkoreit, Jakob and Jones, Llion and Gomez, Aidan N and Kaiser, {\L}ukasz and Polosukhin, Illia},
  journal={NeurIPS},
  year={2017}
}

@article{autoformer,
  title={Autoformer: Decomposition transformers with auto-correlation for long-term series forecasting},
  author={Wu, Haixu and Xu, Jiehui and Wang, Jianmin and Long, Mingsheng},
  journal={NeurIPS},
  year={2021}
}

@inproceedings{fedformer,
  title={Fedformer: Frequency enhanced decomposed transformer for long-term series forecasting},
  author={Zhou, Tian and Ma, Ziqing and Wen, Qingsong and Wang, Xue and Sun, Liang and Jin, Rong},
  booktitle={International conference on machine learning},
  year={2022},
  organization={PMLR}
}

@inproceedings{
patchtst,
title={A Time Series is Worth 64 Words:  Long-term Forecasting with Transformers},
author={Yuqi Nie and Nam H Nguyen and Phanwadee Sinthong and Jayant Kalagnanam},
booktitle={ICLR},
year={2023}
}

@inproceedings{
crossformer,
title={Crossformer: Transformer Utilizing Cross-Dimension Dependency for Multivariate Time Series Forecasting},
author={Yunhao Zhang and Junchi Yan},
booktitle={ICLR},
year={2023}
}

@inproceedings{
itransformer,
title={iTransformer: Inverted Transformers Are Effective for Time Series Forecasting},
author={Yong Liu and Tengge Hu and Haoran Zhang and Haixu Wu and Shiyu Wang and Lintao Ma and Mingsheng Long},
booktitle={ICLR},
year={2024}
}

@inproceedings{
simpletm,
title={Simple{TM}: A Simple Baseline for Multivariate Time Series Forecasting},
author={Hui Chen and Viet Luong and Lopamudra Mukherjee and Vikas Singh},
booktitle={ICLR},
year={2025}
}

@article{waveform, 
title={WaveForM: Graph Enhanced Wavelet Learning for Long Sequence Forecasting of Multivariate Time Series}, 
abstractNote={Multivariate time series (MTS) analysis and forecasting are crucial in many real-world applications, such as smart traffic management and weather forecasting. However, most existing work either focuses on short sequence forecasting or makes predictions predominantly with time domain features, which is not effective at removing noises with irregular frequencies in MTS. Therefore, we propose WaveForM, an end-to-end graph enhanced Wavelet learning framework for long sequence FORecasting of MTS. WaveForM first utilizes Discrete Wavelet Transform (DWT) to represent MTS in the wavelet domain, which captures both frequency and time domain features with a sound theoretical basis. To enable the effective learning in the wavelet domain, we further propose a graph constructor, which learns a global graph to represent the relationships between MTS variables, and graph-enhanced prediction modules, which utilize dilated convolution and graph convolution to capture the correlations between time series and predict the wavelet coefficients at different levels. Extensive experiments on five real-world forecasting datasets show that our model can achieve considerable performance improvement over different prediction lengths against the most competitive baseline of each dataset.}, 
journal={AAAI}, 
author={Yang, Fuhao and Li, Xin and Wang, Min and Zang, Hongyu and Pang, Wei and Wang, Mingzhong}, 
year={2023}, 
month={Jun.}}

@inproceedings{msgnet,
  title={Msgnet: Learning multi-scale inter-series correlations for multivariate time series forecasting},
  author={Cai, Wanlin and Liang, Yuxuan and Liu, Xianggen and Feng, Jianshuai and Wu, Yuankai},
  booktitle={AAAI},
  year={2024}
}

@article{cyclenet,
  title={Cyclenet: Enhancing time series forecasting through modeling periodic patterns},
  author={Lin, Shengsheng and Lin, Weiwei and Hu, Xinyi and Wu, Wentai and Mo, Ruichao and Zhong, Haocheng},
  journal={NeurIPS},
  year={2024}
}

@article{
tsmixer,
title={{TSM}ixer: An All-{MLP} Architecture for Time Series Forecasting},
author={Si-An Chen and Chun-Liang Li and Sercan O Arik and Nathanael Christian Yoder and Tomas Pfister},
journal={TMLR},
year={2023}
}

@inproceedings{timemixer,
	title={TimeMixer: Decomposable Multiscale Mixing for Time Series Forecasting},
	author={Wang, Shiyu and Wu, Haixu and Shi, Xiaoming and Hu, Tengge and Luo, Huakun and Ma, Lintao and Zhang, James Y and ZHOU, JUN},
	booktitle={ICLR},
	year={2024}
}

@inproceedings{wpmixer,
  title={Wpmixer: Efficient multi-resolution mixing for long-term time series forecasting},
  author={Murad, Md Mahmuddun Nabi and Aktukmak, Mehmet and Yilmaz, Yasin},
  booktitle={AAAI},
  year={2025}
}

@article{frets,
  title={Frequency-domain MLPs are more effective learners in time series forecasting},
  author={Yi, Kun and Zhang, Qi and Fan, Wei and Wang, Shoujin and Wang, Pengyang and He, Hui and An, Ning and Lian, Defu and Cao, Longbing and Niu, Zhendong},
  journal={NeurIPS},
  year={2023}
}

@inproceedings{
vit,
title={An Image is Worth 16x16 Words: Transformers for Image Recognition at Scale},
author={Alexey Dosovitskiy and Lucas Beyer and Alexander Kolesnikov and Dirk Weissenborn and Xiaohua Zhai and Thomas Unterthiner and Mostafa Dehghani and Matthias Minderer and Georg Heigold and Sylvain Gelly and Jakob Uszkoreit and Neil Houlsby},
booktitle={ICLR},
year={2021}
}

@inproceedings{tsmixer2,
  title={Tsmixer: Lightweight mlp-mixer model for multivariate time series forecasting},
  author={Ekambaram, Vijay and Jati, Arindam and Nguyen, Nam and Sinthong, Phanwadee and Kalagnanam, Jayant},
  booktitle={ACM SIGKDD},
  year={2023}
}

@inproceedings{swintransformer,
  title={Swin transformer: Hierarchical vision transformer using shifted windows},
  author={Liu, Ze and Lin, Yutong and Cao, Yue and Hu, Han and Wei, Yixuan and Zhang, Zheng and Lin, Stephen and Guo, Baining},
  booktitle={IEEE/CVF},
  year={2021}
}

@Inbook{swt,
author="Nason, G. P.
and Silverman, B. W.",
editor="Antoniadis, Anestis
and Oppenheim, Georges",
title="The Stationary Wavelet Transform and some Statistical Applications",
bookTitle="Wavelets and Statistics",
year="1995",
publisher="Springer New York",
address="New York, NY",
pages="281--299",
isbn="978-1-4612-2544-7"
}

@article{leranableswt,
   title={Fully learnable deep wavelet transform for unsupervised monitoring of high-frequency time series},
   journal={Proceedings of the National Academy of Sciences},
   publisher={Proceedings of the National Academy of Sciences},
   author={Michau, Gabriel and Frusque, Gaetan and Fink, Olga},
   year={2022},
   month=feb }

@article{timexer,
  title={Timexer: Empowering transformers for time series forecasting with exogenous variables},
  author={Wang, Yuxuan and Wu, Haixu and Dong, Jiaxiang and Qin, Guo and Zhang, Haoran and Liu, Yong and Qiu, Yunzhong and Wang, Jianmin and Long, Mingsheng},
  journal={NeurIPS},
  year={2024}
}

@misc{optuna,
      title={Optuna: A Next-generation Hyperparameter Optimization Framework}, 
      author={Takuya Akiba and Shotaro Sano and Toshihiko Yanase and Takeru Ohta and Masanori Koyama},
      year={2019},
      eprint={1907.10902},
      archivePrefix={arXiv},
      primaryClass={cs.LG}
}
\bibliographystyle{splncs04}

\newpage
\appendix
\renewcommand{\theHtable}{A\arabic{table}}
\renewcommand{\theHfigure}{A\arabic{figure}}
\section{Experimental Settings}
\subsection{Datasets}
\begin{table}[htbp]
   \centering
   \small
   \caption{Summary of dataset statistics and experimental configurations.}
   \label{tab:dataset_stats}
   \renewcommand{\arraystretch}{1.2}
   \begin{tabular}{l|c|c|c}
   \toprule
   \multirow{2}{*}{Dataset} & \multicolumn{2}{c|}{Statistics} & Dataset Split \\
   \cmidrule{2-4}
   & $C$ & $T$ & ($N_{\text{train}}, N_{\text{val}}, N_{\text{test}}$) \\
   \midrule
   ETTh1 \cite{informer} & 7 & 17,420 & (8,545, 2,881, 2,881) \\
   ETTh2 \cite{informer} & 7 & 17,420 & (8,545, 2,881, 2,881) \\
   ETTm1 \cite{informer} & 7 & 69,680 & (34,465, 11,521, 11,521) \\
   ETTm2 \cite{informer} & 7 & 69,680 & (34,465, 11,521, 11,521) \\
   Weather \cite{autoformer} & 21 & 52,696 & (36,792, 5,271, 10,540) \\
   ECL \cite{autoformer} & 321 & 26,304 & (18,317, 2,633, 5,261) \\
   Traffic \cite{autoformer} & 862 & 17,544 & (12,185, 1,756, 3,508) \\
   \bottomrule
   \end{tabular}
 \end{table}

 To evaluate the performance of our proposed model, we conduct extensive experiments on seven widely used real-world multivariate time series benchmarks. The statistical details and chronological train-validation-test splits of these datasets are summarized in Table~\ref{tab:dataset_stats}. 

 \begin{itemize}
     \item \textbf{ETT (Electricity Transformer Temperature) \cite{informer}:} This dataset contains data collected from two separate electricity transformers in China, recorded at two different granularities: hourly (\textbf{ETTh1}, \textbf{ETTh2}) and 15-minute intervals (\textbf{ETTm1}, \textbf{ETTm2}). Each dataset consists of 7 features, which include the target variable (oil temperature) and 6 external power load features.
     
     \item \textbf{Weather \cite{autoformer}:} This dataset comprises 21 meteorological indicators (such as temperature, humidity, and visibility) recorded every 10 minutes by the Max Planck Institute for Biogeochemistry weather station in Germany. It exhibits complex seasonal and daily patterns.
     
     \item \textbf{ECL (Electricity Consuming Load) \cite{autoformer}:} This dataset tracks the hourly electricity consumption of 321 different clients. It is widely used to evaluate forecasting models on high-dimensional data with complex temporal dependencies.
     
     \item \textbf{Traffic \cite{autoformer}:} This dataset contains hourly road occupancy rates measured by 862 different sensors placed on freeways across the San Francisco Bay Area. It represents a large-scale forecasting scenario with strong spatial-temporal correlations.
 \end{itemize}
 
 Following standard protocols in regular time series forecasting, we split all datasets chronologically into training, validation, and testing sets to prevent data leakage. Specifically, the ETT datasets are split using a ratio of 6:2:2, while the Weather, ECL, and Traffic datasets follow a 7:1:2 split ratio.
 \subsection{Hyperparameter Optimization}

 \subsubsection{Hyperparameter Space}
 To automatically discover the optimal model configurations, we leverage the Optuna library \cite{optuna}. The search is guided by the Tree-structured Parzen Estimator (TPE) algorithm, which helps efficiently navigate the parameter space while maintaining reproducibility. We independently tune the model for each prediction horizon $T \in \{96, 192, 336, 720\}$, setting the objective to minimize the Mean Squared Error (MSE) on the validation set. 
 
 The complete search space investigated during this tuning phase—covering network dimensions, patch configurations, and training hyperparameters—is outlined in Table~\ref{tab:hyperparams}. For the detailed configurations for each dataset and prediction length, please refer to the GitHub repository linked in Section~\ref{introduction}.
 \begin{table}[htbp]
   \centering
   \small
   \caption{Hyperparameter search space explored using Optuna.}
   \label{tab:hyperparams}
   \renewcommand{\arraystretch}{1.2}
   \begin{tabular}{l|l}
   \toprule
   \textbf{Hyperparameter} & \textbf{Search Space / Values} \\
   \midrule
   Learning Rate & Float $\in [10^{-4}, 10^{-2}]$ (Log-scale) \\
   $d_{\text{model}}$ & $\{32, 64, 128, 256, 512, 1024\}$ \\
   $d_{\text{ff}}$ & $\{32, 64, 128, 256, 512, 1024, 2048\}$ \\
   Dropout & Float $\in [0.4, 0.9]$ \\
   Encoder Layers ($e_{\text{layers}}$) & Integer $\in [1, 5]$ \\
   FC Dropout & $\{0, 0.1, 0.2\}$ \\
   Wavelet Levels ($J$) & Integer $\in [1, 5]$ \\
   Patch Size & $\{4, 8, 12, 16, 24, 32, 48\}$ \\
   \bottomrule
   \end{tabular}
 \end{table}

\section{Efficiency Analysis}

To evaluate the computational efficiency of our proposed architecture, we compare HPMixer against SimpleTM on the Weather and ETTm1 datasets (look-back $L=96$, horizon $T=96$). The results are summarized in Table~\ref{tab:efficiency_comparison_consolidated}. 
While HPMixer incurs a higher parameter count and theoretical computational complexity (GFLOPs) due to the multi-branch design and the Learnable Stationary Wavelet Transform, it achieves a highly favorable trade-off in practical execution. Most notably, HPMixer not only delivers superior predictive accuracy (consistently lower MSE) but also operates at a significantly faster practical training speed (lower seconds per iteration) across both datasets. This acceleration demonstrates that our non-overlapping hierarchical patching and channel-mixing mechanisms are highly parallelizable, resulting in an architecture that is both highly accurate and practically efficient.

\begin{table}[ht]
   \centering
   \caption{Efficiency Analysis: Comparison of HPMixer (Ours) and SimpleTM across Weather and ETTm1 datasets ($L=96, T=96$).}
   \label{tab:efficiency_comparison_consolidated}
   \small
   \begin{tabular}{llcccc}
   \toprule
   \textbf{Dataset} & \textbf{Model} & \textbf{MSE} $\downarrow$ & \textbf{Total Params} & \textbf{GFLOPs} & \textbf{Speed (s/iter)} $\downarrow$ \\ 
   \midrule
   \textbf{Weather} & \textbf{HPMixer (Ours)} & \textbf{0.154} & 172,172 & 0.1741 & \textbf{0.1364} \\
    & SimpleTM & 0.162 & \textbf{14,880} & \textbf{0.0099} & 0.2424 \\
   \midrule
   \textbf{ETTm1} & \textbf{HPMixer (Ours)} & \textbf{0.305} & 5,324,504 & 0.9802 & \textbf{0.2488} \\
    & SimpleTM & 0.321 & \textbf{1,700,576} & \textbf{0.7611} & 0.4210 \\
   \bottomrule
   \end{tabular}
\end{table}

\section{Cycle Length and Patch Size Robustness Analysis}

\begin{figure}[htbp]
   \centering
   \includegraphics[width=0.8\textwidth]{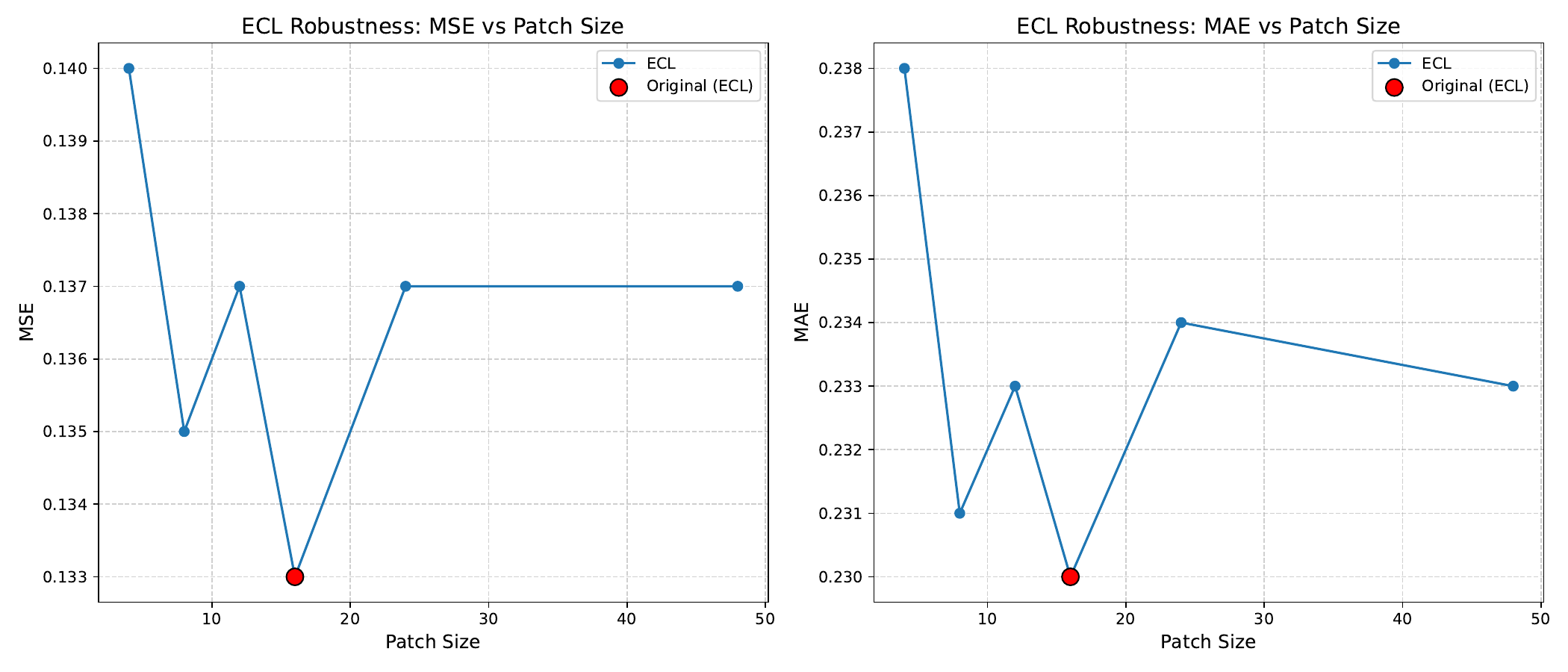}
   \vspace{0.2cm} 
 
   \includegraphics[width=0.8\textwidth]{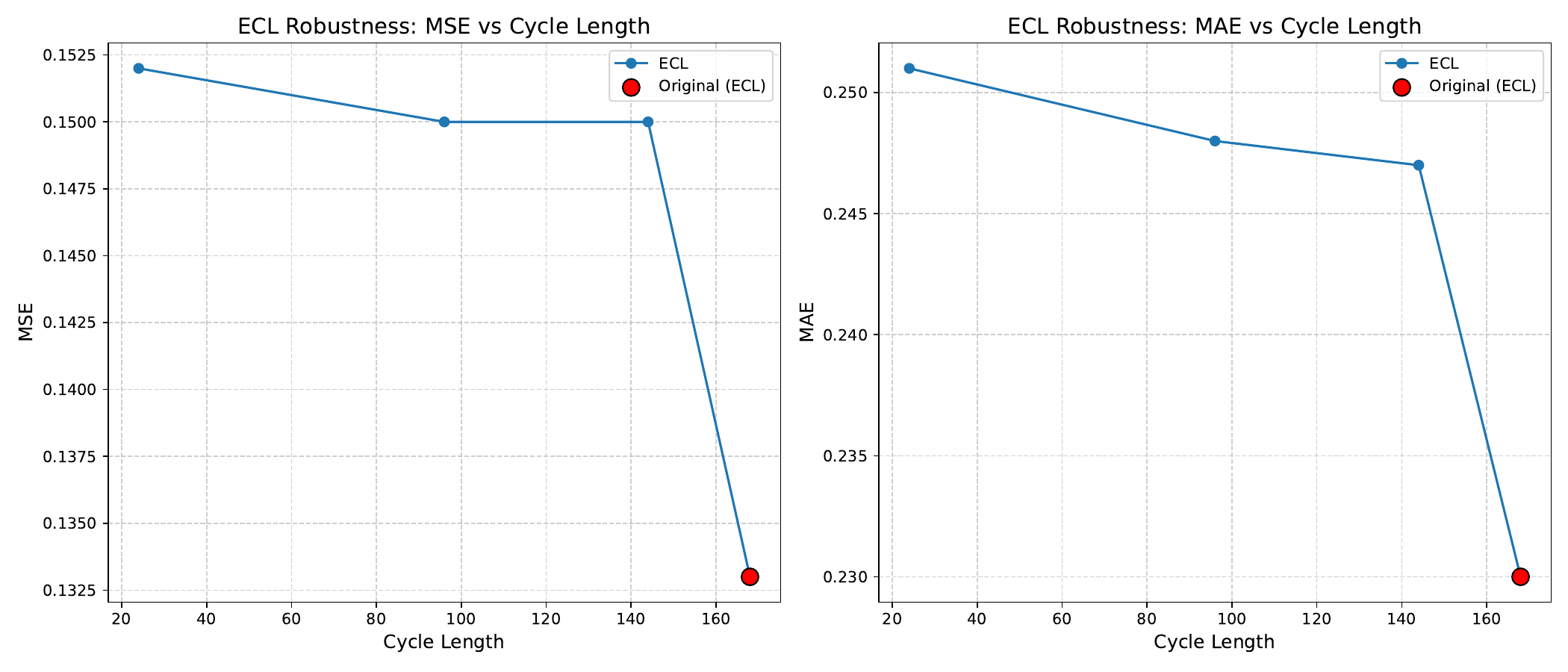}
   \vspace{0.2cm}
 
   \includegraphics[width=0.8\textwidth]{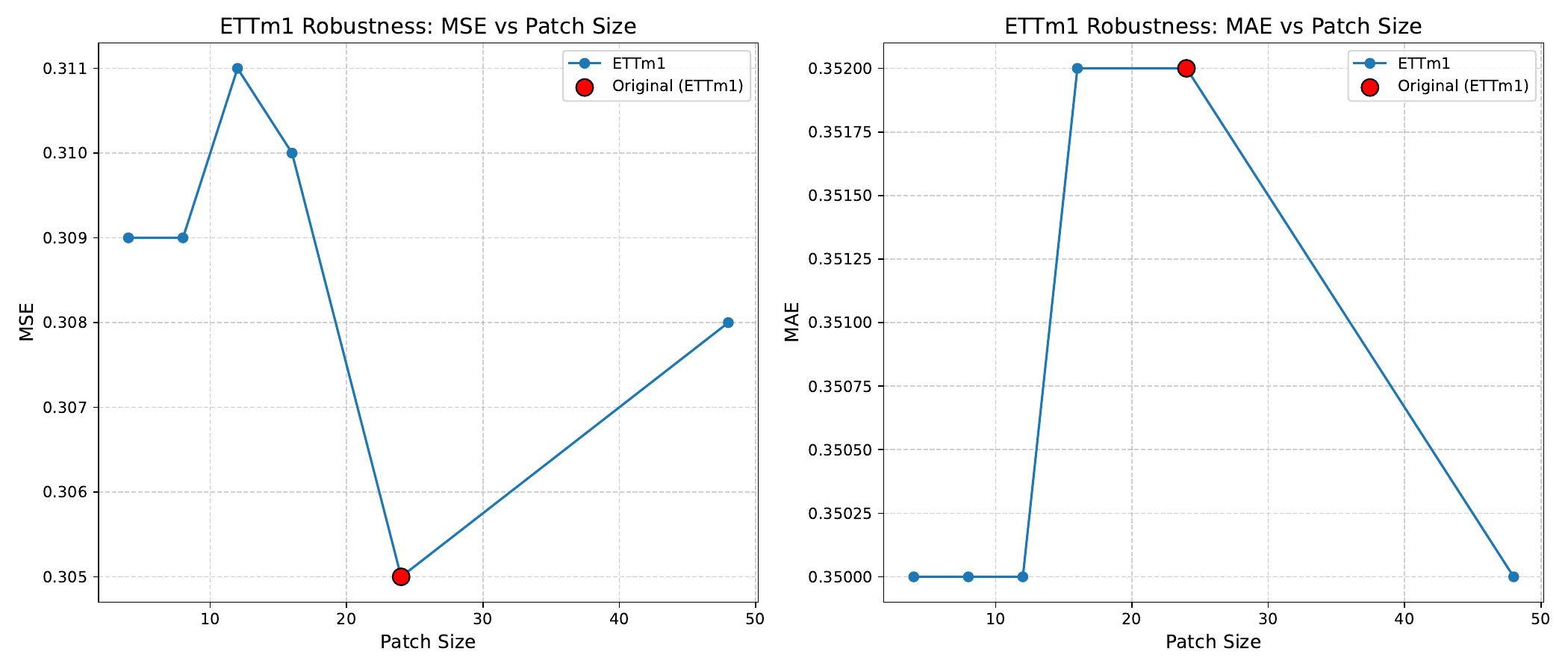}
   \vspace{0.2cm}
 
   \includegraphics[width=0.8\textwidth]{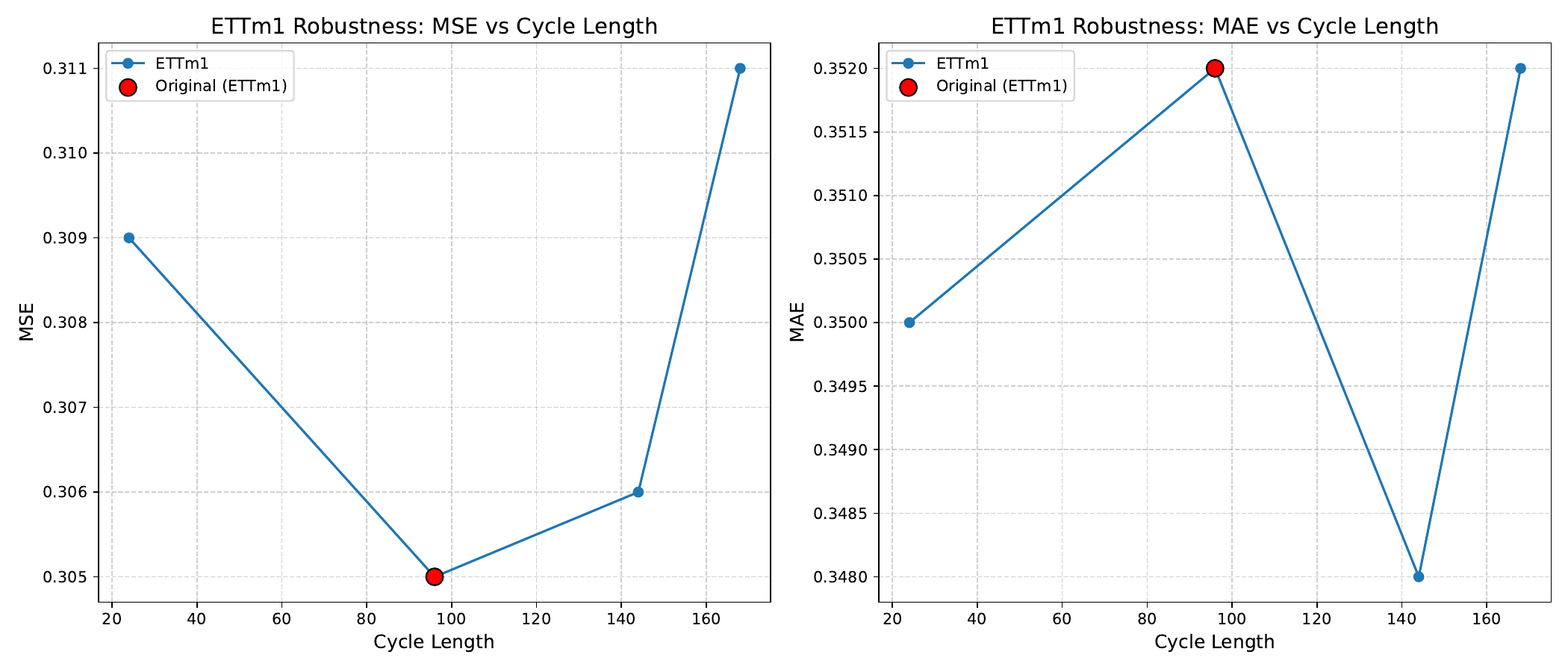}
 
   \caption{Robustness analysis evaluating the impact of varying patch sizes and cycle lengths on the ECL and ETTm1 datasets. The original optimal configurations are marked to demonstrate their superiority in minimizing the Mean Squared Error.}
   \label{fig:robustness_analysis}
 \end{figure}

To evaluate the stability and sensitivity of our proposed model, we conduct a robustness analysis on two critical hyperparameters: cycle length and patch size. The experiments are performed on the ECL and ETTm1 datasets, observing how variations in these parameters impact the Mean Squared Error (MSE) and Mean Absolute Error (MAE).

The model demonstrates strong robustness and a clear performance peak on the ECL dataset. For the patch size variations, our original configuration (patch size 16) achieves the lowest error rates, significantly outperforming alternative sizes. A similar trend is observed in the cycle length analysis, where the original setting of 168 yields the best MSE and MAE. Moving away from these optimal values generally leads to a degradation in performance, confirming that our optimization process successfully identified the precise parameters needed to capture the intrinsic temporal patterns of the ECL dataset.

The ETTm1 dataset exhibits more complex temporal dynamics, resulting in minor performance fluctuations across different configurations. In the patch size analysis, while there are slight variations in the MAE, our original configuration (patch size 24) remains superior by achieving the lowest overall MSE. Because MSE disproportionately penalizes larger forecasting errors, minimizing it is crucial for robust long-term forecasting. Similarly, for the cycle length, the original configuration of 96 maintains the lowest MSE despite minor fluctuations in the MAE across the tested spectrum. 

While some secondary metric fluctuations naturally occur depending on the specific dataset and parameter combination, the overall analysis validates our model's structural robustness. The originally selected hyperparameters consistently deliver the most optimal and balanced forecasting performance, proving that our tuning strategy effectively isolates the most reliable configurations for varying time-series characteristics.

\section{Visual Analysis of Component Decoupling}

\begin{figure}[htbp]
   \centering
   \includegraphics[width=0.9\textwidth]{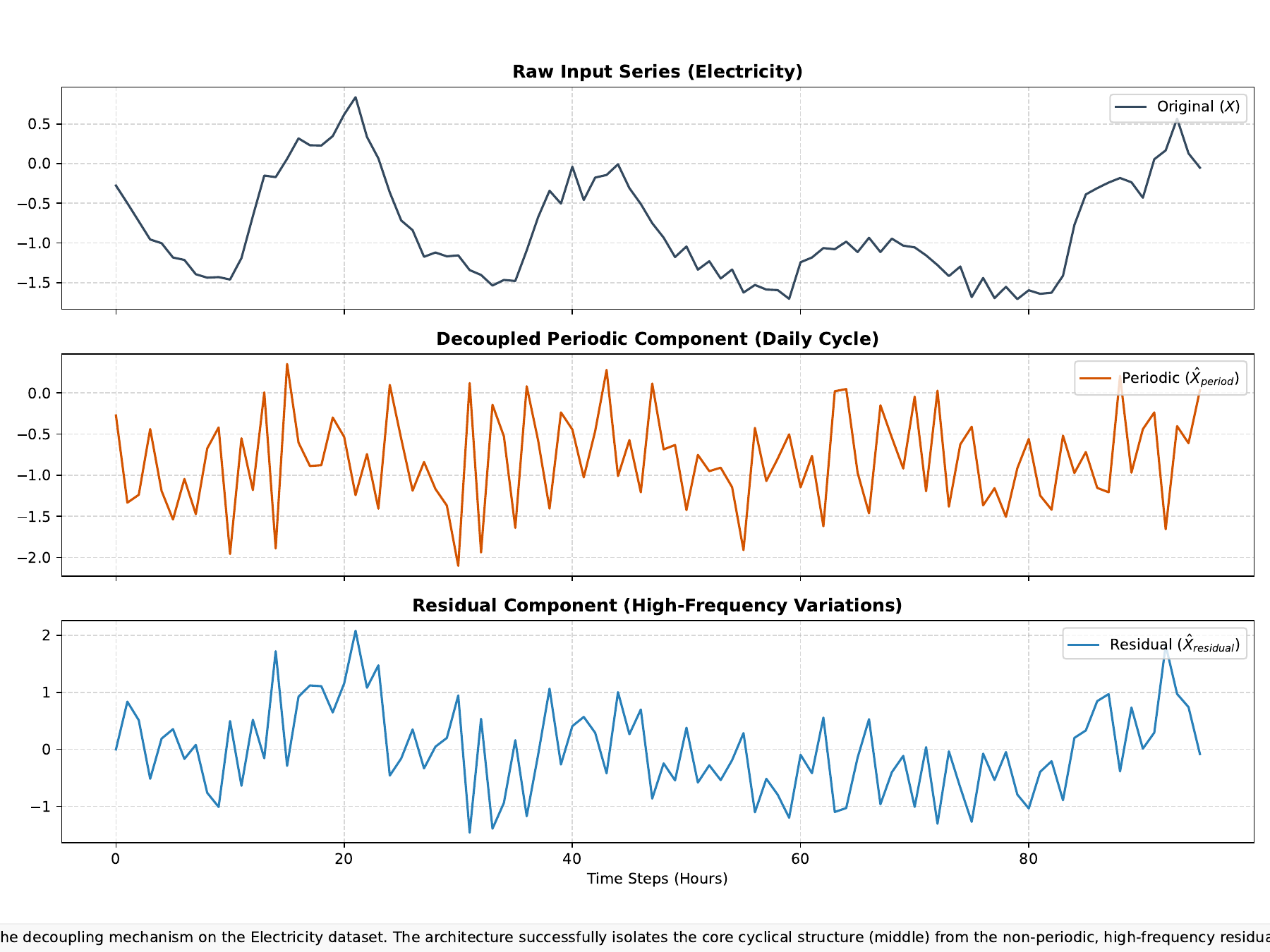}
   
   \vspace{0.2cm} 
   
   \includegraphics[width=0.9\textwidth]{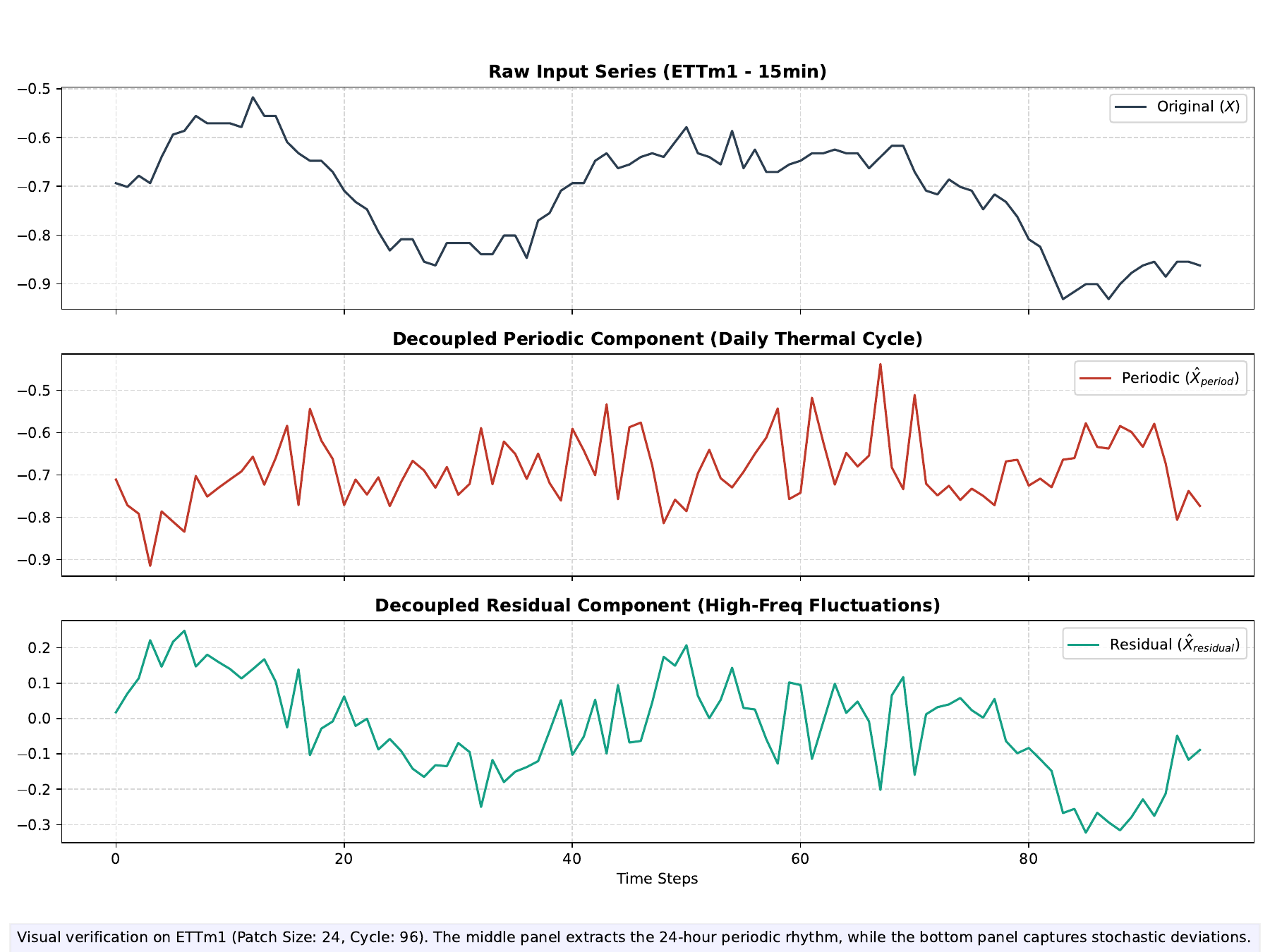}
   
   \caption{Visual verification of the decoupling mechanism on the Electricity (top) and ETTm1 (bottom) datasets. The architecture successfully isolates the core cyclical structure (middle panels) from the non-periodic, high-frequency residuals and stochastic deviations (bottom panels).} 
   \label{fig:decoupling_visuals}
 \end{figure}

To intuitively understand the internal mechanics of our proposed architecture, we visualize the intermediate outputs of the decoupling module. Figure~\ref{fig:decoupling_visuals} demonstrates how the raw input series is decomposed into a periodic (cyclical) component and a residual (trend and high-frequency) component for both the Electricity and ETTm1 datasets \cite{autoformer}\cite{informer}.

As observed in the visualizations, the decoupled residual component effectively captures the broader structural trends and prominent stochastic deviations, closely mirroring the overall trajectory of the original data.

However, our architecture is designed such that these components do not operate in isolation. Ultimately, it is the synergistic combination of the extracted core seasonality (the cycle) and the broader trend variations (the residual) that enables the model to reconstruct the signal and achieve highly accurate, robust forecasting performance.

\end{document}